\documentclass[runningheads]{llncs}
\usepackage[T1]{fontenc}
\usepackage{epsfig}
\usepackage{graphicx}
\usepackage{amsmath}
\usepackage{amssymb}
\usepackage{bm}
\usepackage{subfig}
\usepackage{caption}
\usepackage{diagbox}
\usepackage{multirow}
\usepackage{float}
\usepackage[table]{xcolor}
\usepackage{soul}
\usepackage{balance}
\usepackage{hyperref}[breaklinks=false]
\graphicspath{{IMAGES/}}
\usepackage{balance}

\begin{document}

\title{Can GAN-induced Attribute Manipulations Impact Face Recognition?}
%
%
\author{Sudipta Banerjee\thanks{Corresponding author}\inst{1}\orcidID{0000-0003-1623-7697} \and
Aditi Aggarwal\inst{1} \and
Arun Ross\inst{2}\orcidID{0000-0001-8850-3013}}
\institute{International Institute
of Information Technology,
Hyderabad, India \\
\email{sudipta.b@iiit.ac.in, aditi.aggarwal@students.iiit.ac.in}
\and
Michigan State University,
East Lansing, USA \\
\email{rossarun@cse.msu.edu}}
\maketitle              

\begin{abstract}
Impact due to demographic factors such as age, sex, race, etc., has been studied extensively in automated face recognition systems. However, the impact of \textit{digitally modified} demographic and facial attributes on face recognition is relatively under-explored. In this work, we study the effect of attribute manipulations induced via generative adversarial networks (GANs) on face recognition performance. We conduct experiments on the CelebA dataset by intentionally modifying thirteen attributes using AttGAN and STGAN and evaluating their impact on two deep learning-based face verification methods, ArcFace and VGGFace. Our findings indicate that some attribute manipulations involving eyeglasses and digital alteration of sex cues can significantly impair face recognition by up to 73\% and need further analysis. 
\keywords{Face recognition  \and Generative adversarial network (GAN) \and Attribute manipulation.}
\end{abstract}

\section{Introduction}
Demographic attributes (race, age and sex)~\cite{DemographicsMITRE,NIST2}, face and hair accessories or attributes (glasses, makeup, hairstyle, beard and hair color)~\cite{Terhorst}, and data acquisition factors (environment and sensors)~\cite{Bowyer_Location,Flynn} play an important role in the performance of automated face recognition systems. Demographic factors can potentially introduce biases in face recognition systems and are well studied in the literature~\cite{NIST1,Rattani,Bowyer_2}. Work to mitigate biases due to demographic factors are currently being investigated~\cite{PrivacyNet,DebiasingECCV}. Typically, some of these attributes can be modified \textit{physically}, \textit{e.g.}, by applying hair dye or undergoing surgery. But what if these attributes are \textit{digitally} modified? Individuals can alter their facial features in photos using image editors for cheekbone highlighting, forehead reduction, etc. The modified images, with revised features, may be posted in social media websites. But do such manipulations affect the biometric utility of these images? Assessing the impact of digital retouching on biometric identification accuracy was done in~\cite{Retouching}. With the arrival of generative adversarial networks (GANs), the possibilities of automated attribute editing have exploded~\cite{AttGAN,STGAN,CafeGAN,GAN_Edit}. GANs can be used to change the direction of hair bangs, remove facial hair, change the intensity of tinted eyeglasses and even add facial expressions to face images. With GANs for facial age progression~\cite{Age_GAN}, a person's appearance can seamlessly transit from looking decades younger to appearing as an elderly individual. Note that the user does not have to be a deep learning expert to edit attributes in face images. Several smartphone-based applications have such attribute modifications in the form of filters, \textit{e.g.}, FaceApp~\cite{FaceAPP}. Open-source applications make it easy to modify an image by uploading the image, selecting the attribute to be edited, using a slider to regulate the magnitude of the change, and downloading the edited photo. The entire process can be easily accomplished in under five minutes~\cite{BeautyGlow}. Recently, mask-aware face editing methods have emerged~\cite{Mask}. 

\textbf{Motivation:} Although the intent of using image editing routines stems from personal preferences, they can be misused for obscuring an identity or impersonating another identity. The style transfer networks are typically evaluated from the perspective of visual realism, \textit{i.e.,} how realistic do the generated images look? However, we rarely investigate the impact of such digital manipulations on biometric face recognition. Work has been done to localize the manipulations~\cite{FakeLocator}, estimate the generative model responsible for producing the effects~\cite{GANPrints}, and gauge the robustness of face recognition networks with respect to pose and expression~\cite{BMVC}. But we need to investigate GAN-based attribute manipulations from the perspective of biometric recognition. Studying the influence of digital manipulations of both demographic and facial attributes on face images is pivotal because an individual can use the edited images in identification documents. Therefore, it is imperative to assess the impact of GAN-based attribute manipulations on biometric recognition to evaluate the robustness of existing open-source deep learning-based face recognition systems~\cite{Ross_ICB}. \textbf{Our objective is to conduct an investigative study that examines the impact of attribute editing (thirteen attributes) of face images by AttGAN}~\cite{AttGAN} \textbf{and STGAN}~\cite{STGAN} \textbf{on two popular open-source face matchers, namely, ArcFace}~\cite{ArcFace} \textbf{and VGGFace~\cite{VGGFace}}. 

The remainder of the paper is organized as follows. Section~\ref{Sec:Prop} describes image attribute editing GANs and open-source face recognition networks analyzed in the work. Section~\ref{Sec:Expts} describes the experimental protocols followed in this work. Section~\ref{Sec:Results} reports and analyzes the findings. Section~\ref{Sec:Summ} concludes the paper.

\begin{figure*}[!h]
    \centering
    \subfloat[Male subject]
    {
    \includegraphics[scale=0.422]{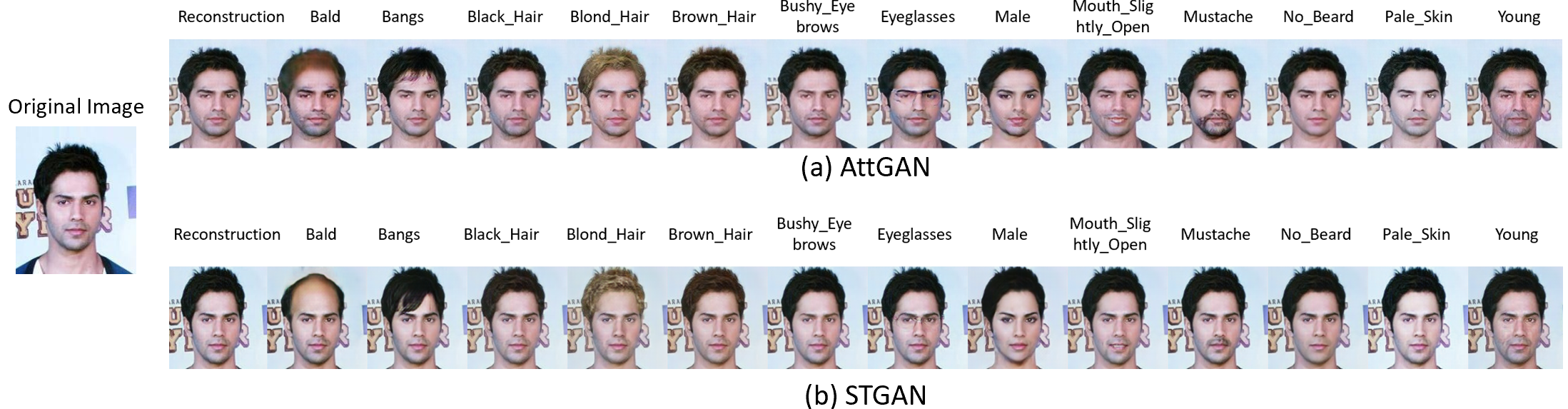}
    }\\
    \subfloat[Female subject]
    {
    \includegraphics[scale=0.42]{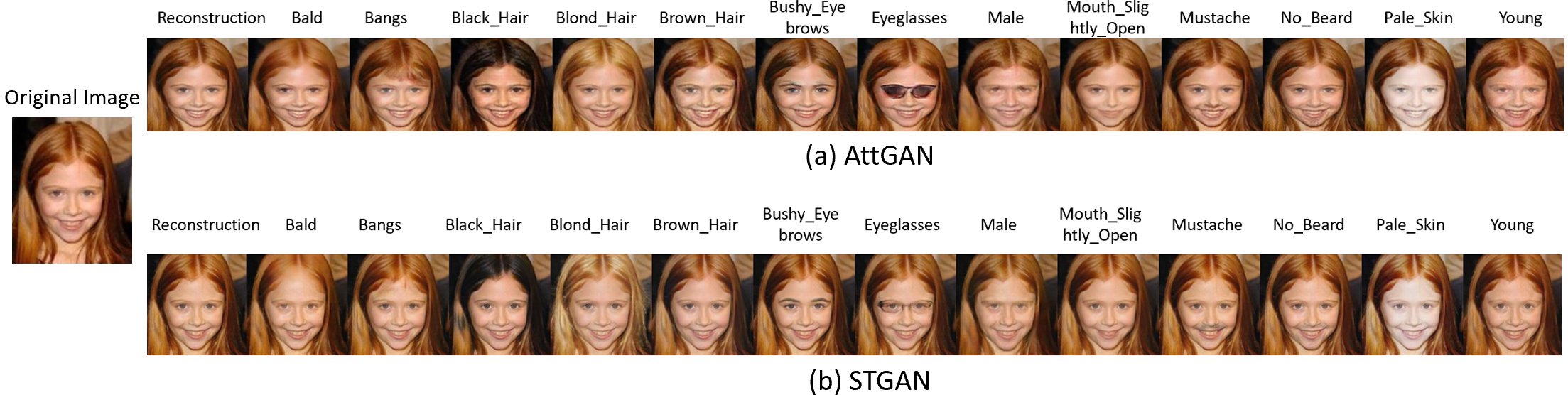}
    }
    \caption{Examples of attribute-manipulated images using AttGAN and STGAN for a (a) male subject and a (b) female subject. To demonstrate the ability of the GAN as a faithful autoencoder, `Reconstruction' images are also displayed.}
    \label{fig:ExampleImgs}
\end{figure*}

\section{Proposed Study} 
\label{Sec:Prop}
In this work, we investigate how GAN-induced attribute manipulations affect face recognition. In the process, we will review the following research questions through our study.
\begin{enumerate}
    \item Does GAN-based attribute editing only produce perceptual changes in face images?
    \item Are there certain attribute manipulations that are more detrimental than others on face verification performance?
    \item Is the impact of GAN-based manipulations consistent across different face recognition networks?
\end{enumerate}

To answer the above questions, we used two GAN-based image editing deep networks, \textit{viz.}, AttGAN and STGAN to modify thirteen attributes on a set of face images from the CelebA dataset. The edited images are then compared with the original face images in terms of biometric utility using two deep learning-based face recognition networks, namely, ArcFace and VGGFace. We hypothesize that digital manipulations induced using GANs can alter the visual perceptibility of the images, but more importantly, affect the biometric utility of the images. Note that these attributes are not originally present in the images: they are artificially induced. Although such manipulations may seem innocuous, they can produce unexpected changes in the face recognition performance and should be handled cautiously. We now discuss the GANs used for attribute manipulation and the deep face networks considered in this work.

\textbf{AttGAN}~\cite{AttGAN}: It can perform binary facial attribute manipulation by modeling the relationship between the attributes and the latent representation of the face. It consists of an encoder, a decoder, an attribute classification network and a discriminator. The attribute classifier imposes an attribute classification constraint to ensure that the generated image possesses the desired attribute. Reconstruction learning is introduced such that the generated image mimics the original image, \textit{i.e.}, only the desired attribute should change without compromising the integrity of the remaining details in the image. Finally, adversarial learning maintains the overall visual realism of the generated images. The network allows high quality facial attribute editing with control over the attribute intensity and accommodates changes in the attribute style.

\textbf{STGAN}~\cite{STGAN}: AttGAN employs an encoder-decoder structure with spatial pooling or downsampling that results in reduction in the resolution and irrecoverable loss of fine details. Therefore, the generated images are susceptible to loss of features and look blurred. Skip connections are introduced in AttGAN that directly concatenate encoder and decoder features and can improve the reconstruction image quality but at the expense of reduced attribute manipulation ability. In contrast, STGAN adopts selective transfer units that can adaptively transform the encoder features supervised by the attributes to be edited. STGAN accepts the difference between the target and source attribute vector, known as the difference attribute vector, as input unlike the AttGAN that takes the entire target attribute vector as input. Taking the difference vector results in more controlled manipulation of the attribute and simplifies the training process. 

\textbf{VGGFace}~\cite{VGGFace}: ``Very deep'' convolutional neural networks are bootstrapped to learn a \textit{N}-way face classifier to recognize \textit{N} subjects. It is designed such that the network associates each training image to a score vector by using the final fully connected layer that comprises \textit{N} linear predictors. Each image is mapped to one of the \textit{N} identities present during training. The score vector needs fine-tuning to perform verification by comparing face descriptors learned in the Euclidean space. To that end, triplet-loss is employed during training to learn a compact face representation (projection) that is well separated across disjoint identities. 

\textbf{ArcFace}~\cite{ArcFace}: Existing face recognition methods attempt to directly learn the face embedding using a softmax loss function or a triplet loss function. However, each of these embedding approaches suffer from drawbacks. Although softmax loss is effective on closed-set face recognition, the face representations are not discriminative enough for open-set face recognition. Triplet loss can cause a huge number of face triplets for large datasets, and semi-hard triplet mining is challenging for effective training. In contrast, additive angular margin loss (ArcFace) optimises the geodesic distance margin by establishing correspondence between the angle and arc in normalized hypersphere. Due to its concise geometric interpretation, ArcFace can provide highly discriminative features for face recognition, and has become one of the most widely-used open-source deep face recognition networks.
\begin{figure*}[!h]
    \subfloat[STGAN-VGGFace]
    {
    \includegraphics[scale=0.25]{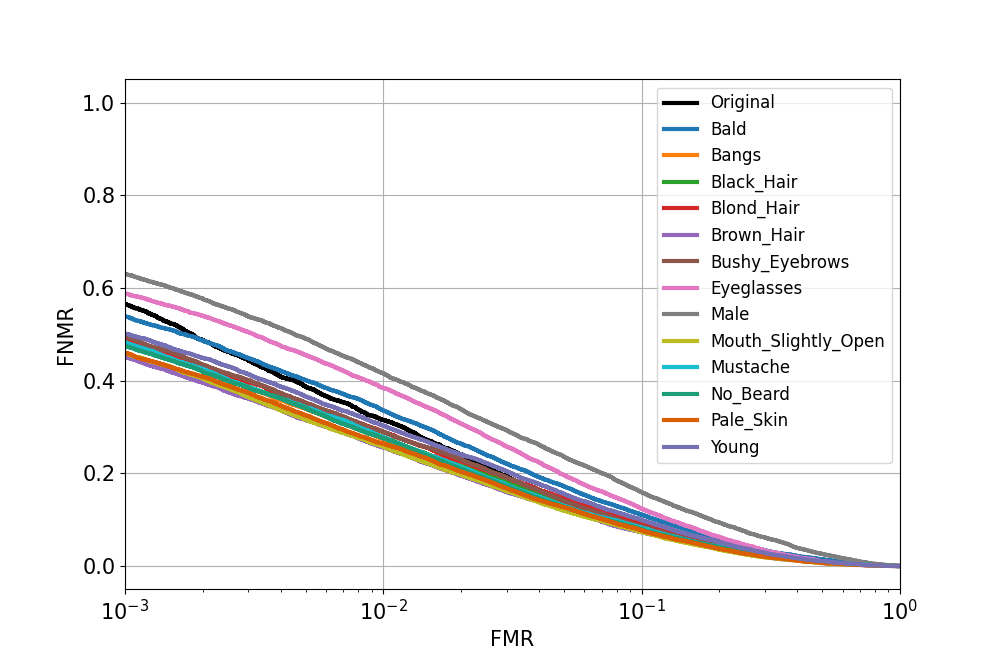}
    }\hspace{-0.65cm}
    \subfloat[STGAN-ArcFace]
    {
    \includegraphics[scale=0.25]{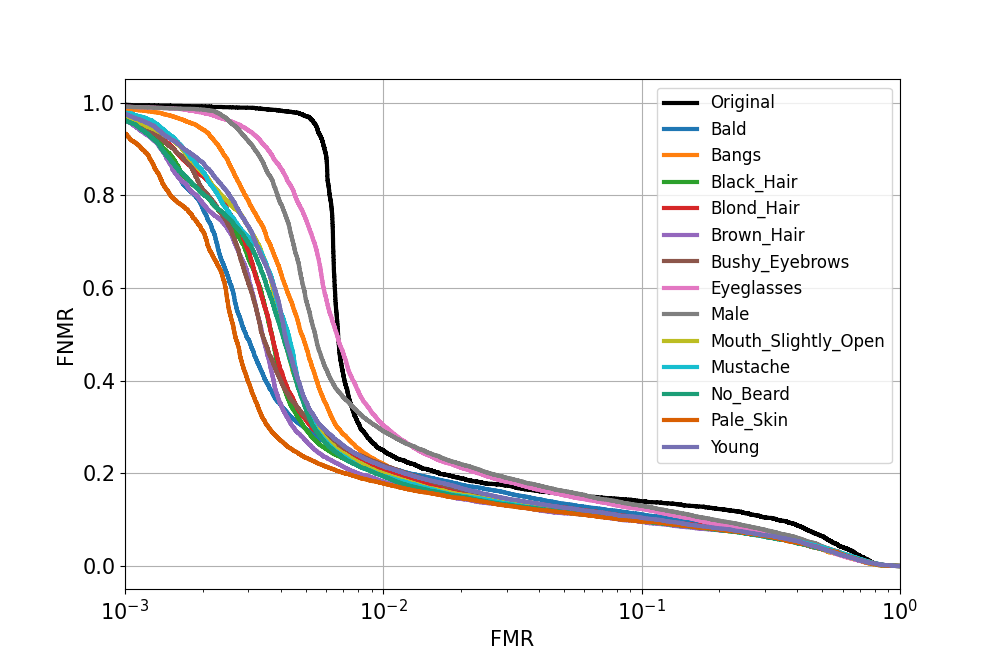}
    }\\
    \subfloat[AttGAN-VGGFace]
    {
    \includegraphics[scale=0.25]{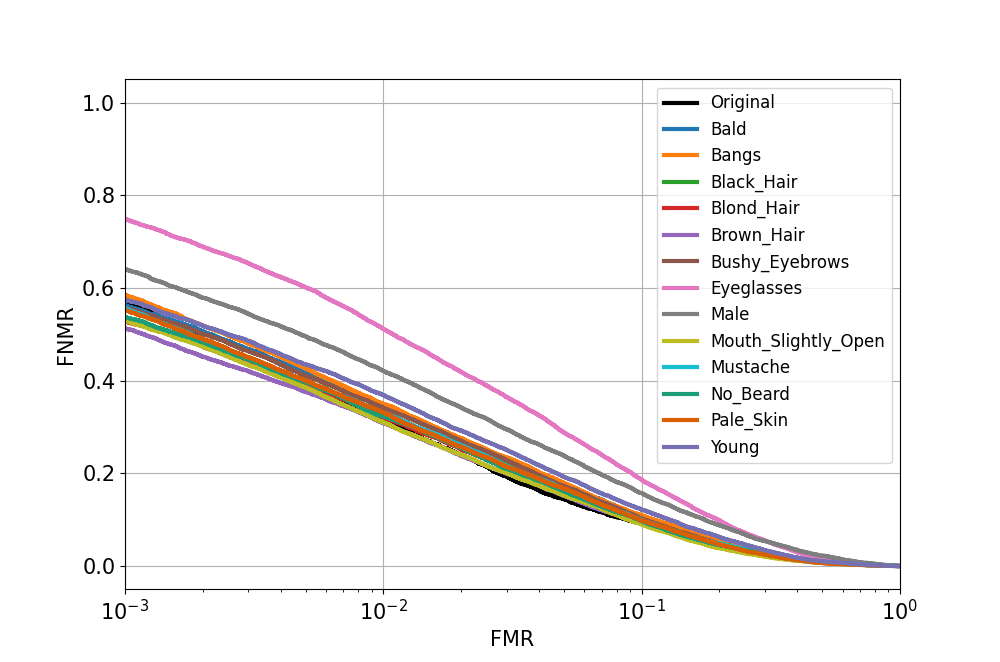}
    }\hspace{-0.65cm}
    \subfloat[AttGAN-ArcFace]
    {
    \includegraphics[scale=0.25]{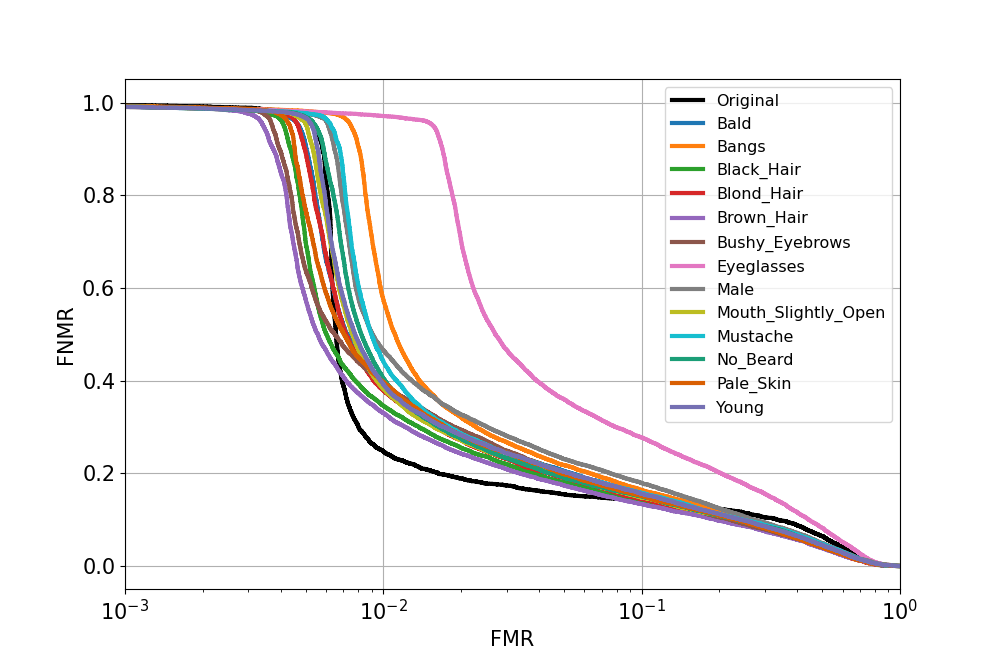}
    }
    \caption{DET curves obtained for all pairs of GANs (for attribute editing) and deep face recognition networks (for biometric matching). (a) STGAN-VGGFace, (b) STGAN-ArcFace, (c) AttGAN-VGGFace and (d) AttGAN-ArcFace.}
    \label{fig:DET_ALL}
\end{figure*}

\section{Experiments} 
\label{Sec:Expts}

We used 2,641 images belonging to 853 unique individuals with an average of $\sim$3 images per subject from the CelebA~\cite{CelebA} dataset for conducting the investigative study in this work. Next, we edited each of these images to alter 13 attributes, one at a time, using AttGAN and STGAN to generate 2,641 $\times$ 13 $\times$ 2 = 68,666 manipulated images. We used the codes and pre-trained models provided by the original authors of AttGAN~\cite{AttGAN_Github} and STGAN~\cite{STGAN_Github} in our work. The list of attributes edited are as follows: \textit{Bald, Bangs, Black\_Hair, Blond\_Hair, Brown\_Hair, Bushy\_Eyebrows, Eyeglasses, Male, Mouth\_Slightly\_Open, Mustache, No\_Beard, Pale\_Skin,} and \textit{Young}. We selected these thirteen attributes as they were modified using the original AttGAN and STGAN models. Examples of GAN edited images are presented in Figures~\ref{fig:ExampleImgs}(a) and \ref{fig:ExampleImgs}(b). Note that the attributes are applied in a toggle fashion. For example, the \textit{Male} attribute changes the face image of the male individual by inducing makeup to impart a feminine appearance (see Figure~\ref{fig:ExampleImgs}(a) ninth image from the left), while the same attribute induces facial hair in the face image of the female subject to impart a masculine appearance (see Figure~\ref{fig:ExampleImgs}(b) ninth image from the left). We can also observe that the manipulations are not identical across the two GANs and can manifest differently across different sexes. For example, attribute \textit{Bald} is more effectively induced by STGAN compared to AttGAN and has a more pronounced effect on male images than female images (see Figures~\ref{fig:ExampleImgs}(a) and (b) second image from the left). 
To perform quantitative evaluation, we compared an original image ($\bm{I_i}$) with its respective attribute manipulated image ($\bm{O_j}$) using a biometric comparator, $\mathcal{B}(\bm{I_i},\bm{O_j})$, where $\mathcal{B}(\cdot,\cdot)$ extracts face representations from each image and computes the vector distance between them. We used cosine distance in this work. The distance value is termed as the biometric match score between a pair of face images. The subscripts indicate the subject identifier. If $i = j$, then $(\bm{I_i},\bm{O_j})$ forms a genuine pair (images belong to the same individual). If $i \neq j$, then $(\bm{I_i},\bm{O_j})$ constitutes an impostor pair (images belong to different individuals). We used the genuine and impostor scores to compute the detection error trade-off (DET) curve, and compared the face recognition performance between the original images and the corresponding attribute manipulated images. In the DET curve, we plot False Non-Match Rate (FNMR) vs. False Match Rate (FMR) at various thresholds. We repeated this process to obtain fourteen curves,\textit{ viz.}, one curve corresponding to the original images (original-original comparison) and the remaining thirteen curves corresponding to the thirteen attributes manipulated by the GAN (original-attribute edited comparison). We used an open-source library for the implementation of ArcFace and VGGFace~\cite{DeepFace_Github}.

\begin{figure}[t]
    \subfloat[50K impostor scores]
    {
    \includegraphics[scale=0.25]{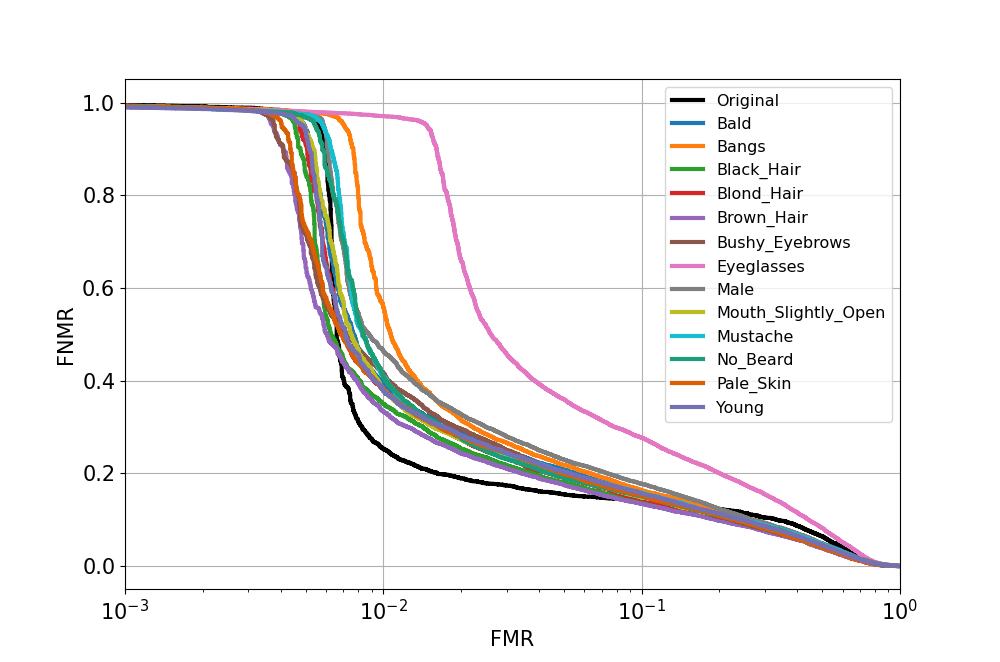}
    } \hspace{-0.65cm} 
    \subfloat[100K impostor scores]
    {
    \includegraphics[scale=0.25]{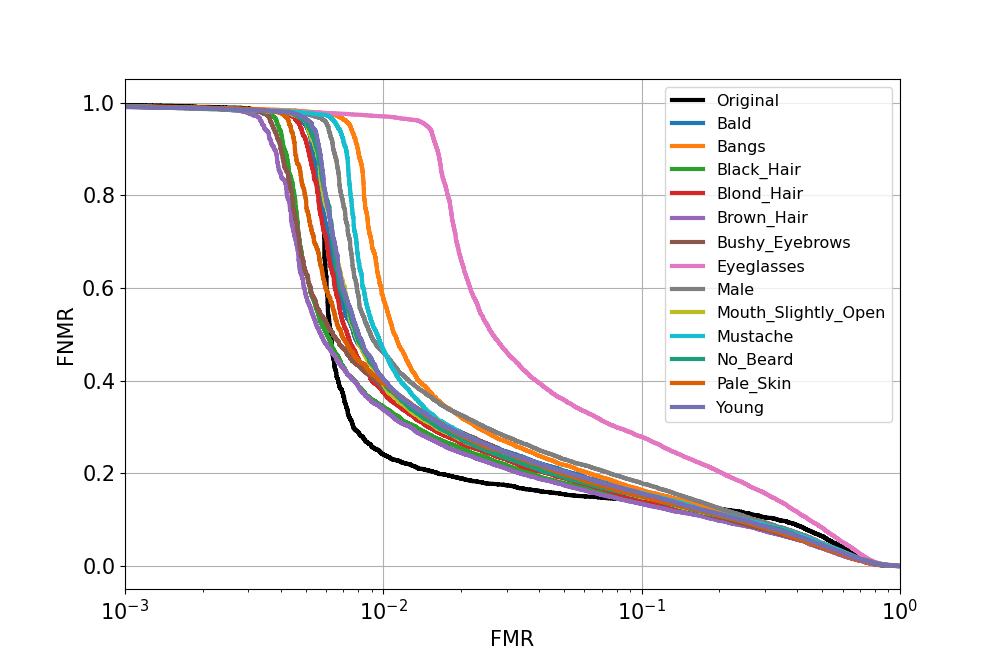}
    } \\
    \subfloat[500K impostor scores]
    {
    \includegraphics[scale=0.25]{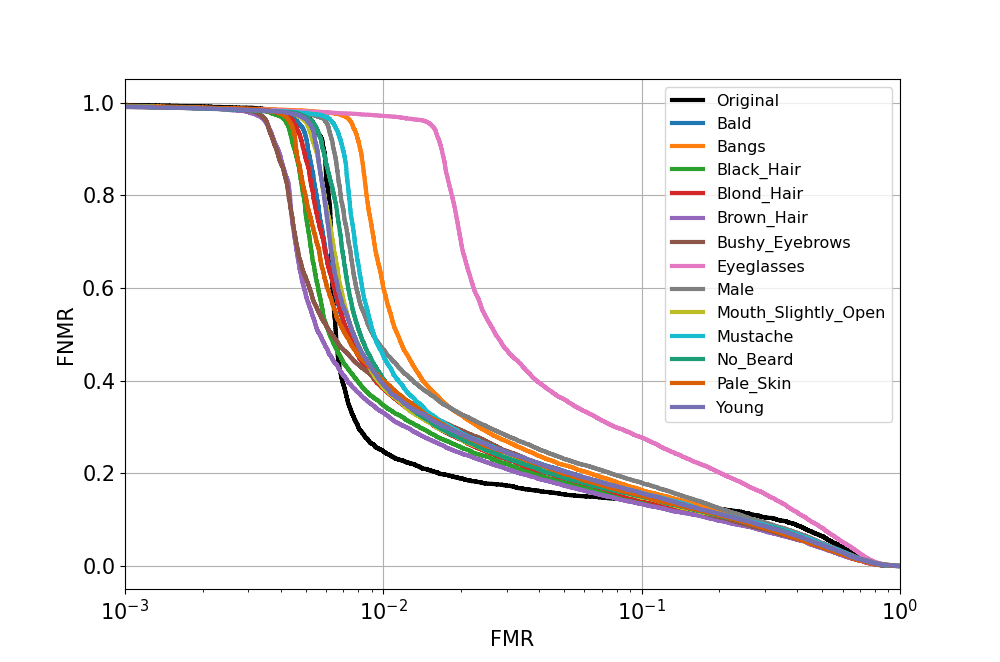}
    } \hspace{-0.65cm} 
    \subfloat[1M impostor scores]
    {
    \includegraphics[scale=0.25]{attgan-arcface_1M.png}
    }
    \caption{DET curves of AttGAN-ArcFace obtained after varying the number of impostor scores (a) 50K, (b) 100K, (c) 500K and (d) 1M. We varied the number of impostor scores with respect to the number of genuine scores ($\sim$10K) in factors of 5X, 10X, 50X and 100X, respectively. Note the DET curves show minimal variations.}
    \label{fig:DETimpostor}
\end{figure}

\section{Findings} 
\label{Sec:Results}
We present the DET curves for four different combinations of attribute manipulating GANs and face comparators used in this work, namely, \textbf{STGAN-VGGFace}, \textbf{STGAN-ArcFace}, \textbf{AttGAN-VGGFace} and \textbf{AttGAN-ArcFace} in Figure~\ref{fig:DET_ALL}. We also tabulated the results from the DET curves at FMR=0.01 in Table~\ref{Tab1} and at FMR=0.1 in Table~\ref{Tab2}. 

Additionally, we performed an experiment to determine how the number of impostor scores affect the overall biometric recognition performance for GAN-edited images. The number of scores corresponding to genuine pairs was $\sim$10K, and the number of scores corresponding to impostor pairs was $\sim$7M. Due to memory constraints, we selected a subset of impostor scores for plotting the DET curves. We randomly selected impostor scores, without replacement, in factors of 5X, 10X, 50X and 100X of the total number of genuine scores (10K). This resulted in 50K, 100K, 500K and 1M impostor scores, respectively. We observed that the DET curves obtained by varying the number of impostor scores from the AttGAN-ArcFace combination are almost identical. See Figure~\ref{fig:DETimpostor}. Therefore, for the remaining three combinations, we used the entire set of genuine scores but restricted to utilizing one million impostor scores.

\begin{table}[!h]
\centering
\caption{Face recognition performance: False Non-Match Rate (FNMR) at a False Match Rate \textbf{(FMR) = 0.01} for attribute manipulated images. The red colored cells correspond to the maximum degradation (if any) in the biometric recognition performance compared to the original in each column. The green colored cells correspond to the maximum improvement (if any) in the biometric recognition performance compared to the original in each column. Identical performance results in multiple colored cells within a column. Note majority of the attributes resulted in degradation. The attribute `Eyeglasses' caused the worst degradation in biometric recognition performance followed by the `Male' attribute.}
\scalebox{0.935}{
\begin{tabular}{|l|l|l|l|l|} \hline
                      & \textbf{\begin{tabular}[c]{@{}l@{}}STGAN-\\ VGGFace\end{tabular}} & \textbf{\begin{tabular}[c]{@{}l@{}}STGAN-\\ ArcFace\end{tabular}} & \textbf{\begin{tabular}[c]{@{}l@{}}AttGAN-\\ VGGFace\end{tabular}} & \textbf{\begin{tabular}[c]{@{}l@{}}AttGAN-\\ ArcFace\end{tabular}} \\ \hline \hline
\textcolor{red}{Original}              & 0.32                                                                  &       0.25                                                            &   0.32                                                                 &        0.25                                                            \\ \hline
Bald                  &    0.34                                                              & 0.22                                                                  &       0.35                                                             &       0.40                                                             \\
Bangs                 &     0.28                                                              & 0.23                                                                  &   0.35                                                                 &     0.57                                                               \\
Black\_Hair           &       0.29                                                            & 0.20                                                                  &    0.33                                                                &       0.35                                                             \\
Blond\_Hair           &    0.30                                                               & 0.21                                                                  & 0.33                                                                    &      0.39                                                              \\
Brown\_hair           &    \cellcolor{green!25}0.26                                                               & 0.19                                                                  &         \cellcolor{green!25}0.31                                                           &    0.34                                                                \\
Bushy\_Eyebrows       &      0.30                                                             & 0.22                                                                  &    0.35                                                                &     0.40                                                               \\
Eyeglasses            &      0.39                                                             & \cellcolor{red!25}0.31                                                                  &    \cellcolor{red!25}0.52                                                                &      \cellcolor{red!25}0.98                                                              \\
Male                  &   \cellcolor{red!25}0.42                                                                & 0.30                                                                  &        0.43                                                            &   0.47                                                                 \\
Mouth\_Slightly\_Open &    \cellcolor{green!25}0.26                                                               & 0.21                                                                  &   0.32                                                                 &      0.38                                                              \\
Mustache              &   0.28                                                                & 0.20                                                                  &       0.34                                                             &  0.45                                                                  \\
No\_Beard             &     0.28                                                              & 0.20                                                                  &          0.33                                                          &    0.41                                                                \\
Pale\_Skin            &       0.27                                                            & \cellcolor{green!25}0.18                                                                  &  0.34                                                                  &  0.40                                                                  \\
Young                 &    0.31                                                               & 0.22                                                                  &      0.37                                                              &   0.40         \\ \hline                                                       
\end{tabular}}
\label{Tab1}
\end{table}

\begin{table}[!h]
\centering
\caption{Face recognition performance: False Non-Match Rate (FNMR) at a False Match Rate \textbf{(FMR) = 0.1} for attribute manipulated images. The red colored cells correspond to the maximum degradation (if any) in the biometric recognition performance compared to the original in each column. The green colored cells correspond to the maximum improvement (if any) in the biometric recognition performance compared to the original in each column. Identical performance results in multiple colored cells within a column. Note @FMR=0.1, STGAN-ArcFace (attribute editing using STGAN and face matching using ArcFace) caused the least degradation in face recognition.}
\scalebox{0.935}{
\begin{tabular}{|l|l|l|l|l|} \hline
                      & \textbf{\begin{tabular}[c]{@{}l@{}}STGAN-\\ VGGFace\end{tabular}} & \textbf{\begin{tabular}[c]{@{}l@{}}STGAN-\\ ArcFace\end{tabular}} & \textbf{\begin{tabular}[c]{@{}l@{}}AttGAN-\\ VGGFace\end{tabular}} & \textbf{\begin{tabular}[c]{@{}l@{}}AttGAN-\\ ArcFace\end{tabular}} \\ \hline \hline
\textcolor{red}{Original}              & 0.10                                                                  &  0.14                                                                 &  0.10                                                                  &    0.14                                                                \\ \hline
Bald                  &  0.12                                                                 & 0.12                                                                  &      0.11                                                              &   0.16                                                                 \\
Bangs                 &  0.09                                                                 & 0.11                                                                  &       0.11                                                             &     0.17                                                               \\
Black\_Hair           &  0.09                                                                 & \cellcolor{green!25}0.10                                                                  &    0.11                                                                &   0.14                                                                \\
Blond\_Hair           &    0.10                                                               & 0.11                                                                  &  0.10                                                                  &  0.15                                                                  \\
Brown\_hair           &   \cellcolor{green!25}0.08                                                                & \cellcolor{green!25}0.10                                                                  &       0.10                                                             &     0.14                                                               \\
Bushy\_Eyebrows       &    0.09                                                               & 0.11                                                                  &   0.11                                                                 & 0.16                                                                   \\
Eyeglasses            &   0.13                                                                & 0.13                                                                  &      \cellcolor{red!25}0.19                                                              &     \cellcolor{red!25}0.28                                                               \\
Male                  &   \cellcolor{red!25}0.17                                                                & 0.14                                                                  &     0.16                                                               &           0.18                                                         \\
Mouth\_Slightly\_Open &      \cellcolor{green!25}0.08                                                             & \cellcolor{green!25}0.10                                                                  &     0.10                                                               &   0.15                                                                 \\
Mustache              &     0.09                                                              & \cellcolor{green!25}0.10                                                                  &    0.10                                                                &           0.16                                                         \\
No\_Beard             &    \cellcolor{green!25}0.08                                                               & \cellcolor{green!25}0.10                                                                  &  0.10                                                                  &   0.16                                                                 \\
Pale\_Skin            &     \cellcolor{green!25}0.08                                                              & \cellcolor{green!25}0.10                                                                  &          0.10                                                          &      0.16                                                              \\
Young                 &   0.10                                                                & 0.11                                                                  &     0.13                                                               &   0.16         \\ \hline                                                       
\end{tabular}}
\label{Tab2}
\end{table}

\textbf{Analysis:} In Table~\ref{Tab1}, we reported the results @FMR=0.01. We observed that the attribute `Eyeglasses', a \textit{facial} attribute produced the \textbf{highest} degradation in the biometric recognition performance (see the red colored cells). AttGAN-ArcFace achieved FNMR=0.98 compared to the original performance of FNMR=0.25 @FMR=0.01 (reduction in performance by 73\%). Note that we reported the reduction as the difference between the FNMR obtained for modified images and the original images. The DET curve indicated as \textit{Original} in Figure~\ref{fig:DET_ALL} denotes the face recognition performance computed using only the unmodified images. AttGAN-VGGFace achieved FNMR=0.52 compared to the original performance of FNMR=0.32 @FMR=0.01 (reduction in performance by 20\%). STGAN-ArcFace achieved FNMR=0.31 compared to the original performance of FNMR=0.25 @FMR=0.01 (reduction in performance by 6\%). STGAN-VGGFace achieved FNMR=0.39 compared to the original performance of FNMR=0.32 @FMR=0.01 (reduction in performance by 7\%). The \textbf{second highest} degradation was caused due to the change in `Male', a \textit{demographic} attribute. AttGAN-ArcFace achieved FNMR=0.47 compared to the original performance of FNMR=0.25 @FMR=0.01 (reduction in performance by 22\%). AttGAN-VGGFace achieved FNMR=0.43 compared to the original performance of FNMR=0.32 @FMR=0.01 (reduction in performance by 11\%). STGAN-ArcFace achieved FNMR=0.30 compared to the original performance of FNMR=0.25 @FMR=0.01 (reduction in performance by 5\%). STGAN-VGGFace achieved FNMR=0.42 compared to the original performance of FNMR=0.32 @FMR=0.01 (reduction in performance by 10\%). STGAN-ArcFace achieved FNMR=0.18 compared to the original performance of FNMR=0.25 @FMR=0.01 (improvement in performance by 7\%) for the attribute `Pale\_Skin'.
In Table~\ref{Tab2}, we reported the results @FMR=0.1. We observed that the attributes seem to improve the recognition performance by up to 4\% in FNMR for STGAN-ArcFace. On the contrary, for the remaining three sets, \textit{viz.}, STGAN-VGGFace, AttGAN-ArcFace and AttGAN-VGGFace, we continued to observe a drop in recognition performance with `Eyeglasses' resulting in the worst drop in performance by 14\%, followed by `Male' resulting in the second highest drop in performance by 7\%.
Now, let us review the questions posited at the beginning of the paper.
\par
 \textit{Question \#1:} Does GAN-based attribute editing only produce perceptual changes in face images?\\ \textit{Observation:} \textbf{No.} GAN-based attribute editing not only alters the perceptual quality of the images but also significantly impacts the biometric recognition performance. See Tables~\ref{Tab1} and~\ref{Tab2}, where a majority of the attributes strongly degraded the face recognition performance.
\par
 \textit{Question \#2:} Are there certain attribute manipulations that are more detrimental than others on face verification performance? \\ \textit{Observation:} \textbf{Yes.} Editing of `Eyeglasses' attribute caused a significant degradation in FNMR by up to 73\% @FMR=0.01 for AttGAN-ArcFace. It was followed by the the `Male' attribute that caused the second highest degradation in FNMR by up to 22\% @FMR=0.01 for AttGAN-ArcFace. Surprisingly, a facial attribute, `Eyeglasses' and a demographic attribute, `Male', when edited separately, were responsible for significant degradation in the biometric recognition performance in a majority of the scenarios. 
\par
\textit{Question \#3:} Is the impact of GAN-based manipulations consistent across different face recognition networks? \\ \textit{Observation:} \textbf{No.} The impact of GAN-based attribute manipulations on face recognition depends on two factors, firstly, which GAN was used to perform the attribute editing operation, and secondly, which face recognition network was used to measure the biometric recognition performance. For example, the `Bald' attribute caused degradation in the performance for AttGAN, \textit{irrespective} of which face matcher was used. However, the same attribute caused improvement in the face recognition accuracy for STGAN when ArcFace was used as the face matcher, but caused reduction in the recognition performance when VGGFace was used as the face comparator. Similar findings were observed for the `Bangs' attribute.


\begin{figure}[h]
    
    \centering
    \includegraphics[scale=0.35]{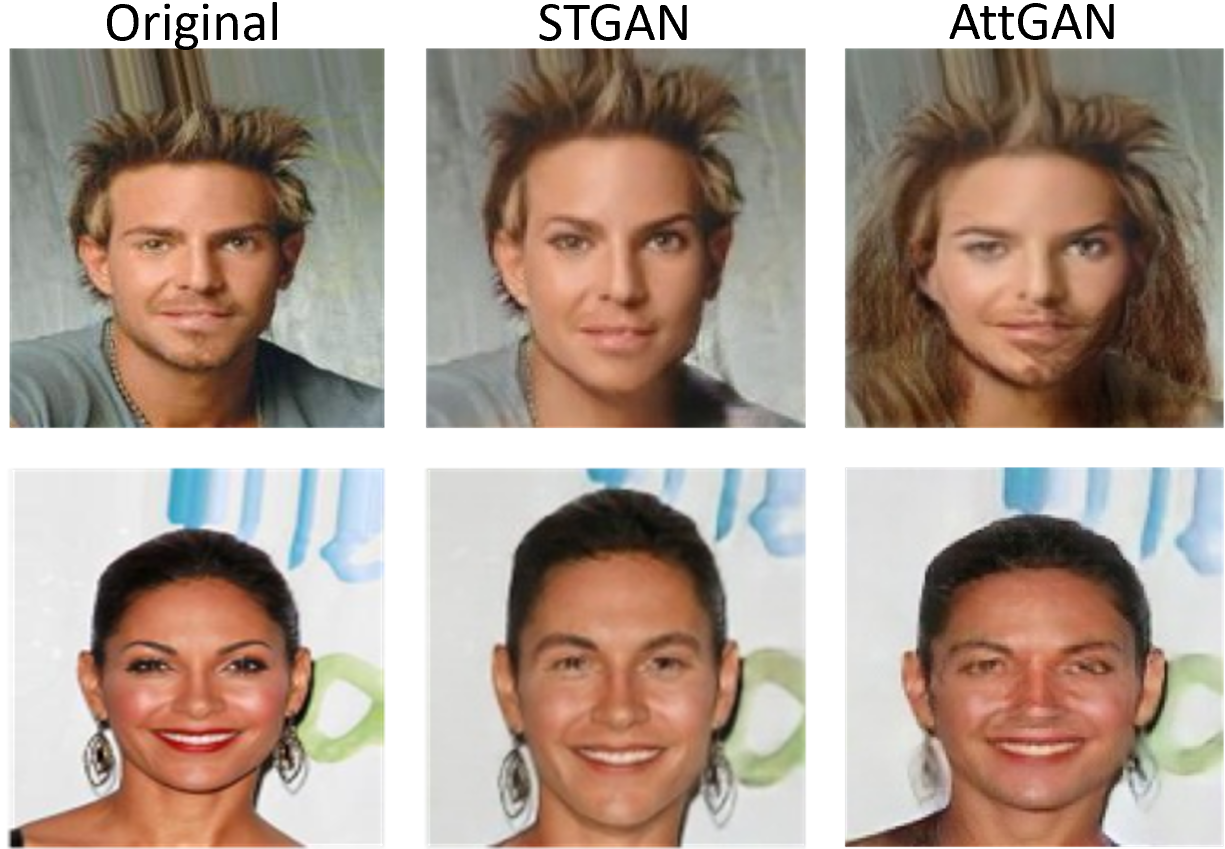}
    
    \caption{Examples depicting `Male' attribute edited using STGAN and AttGAN, respectively, in male (top row) and female (bottom row) individuals. AttGAN output images appear to contain more visual artifacts compared to STGAN outputs.}
    \label{fig:MaleAttrib}
\end{figure}

\begin{figure}[h]
    \centering
    \subfloat[Without Eyeglasses]
    {
    \centering
    \includegraphics[scale=0.25]{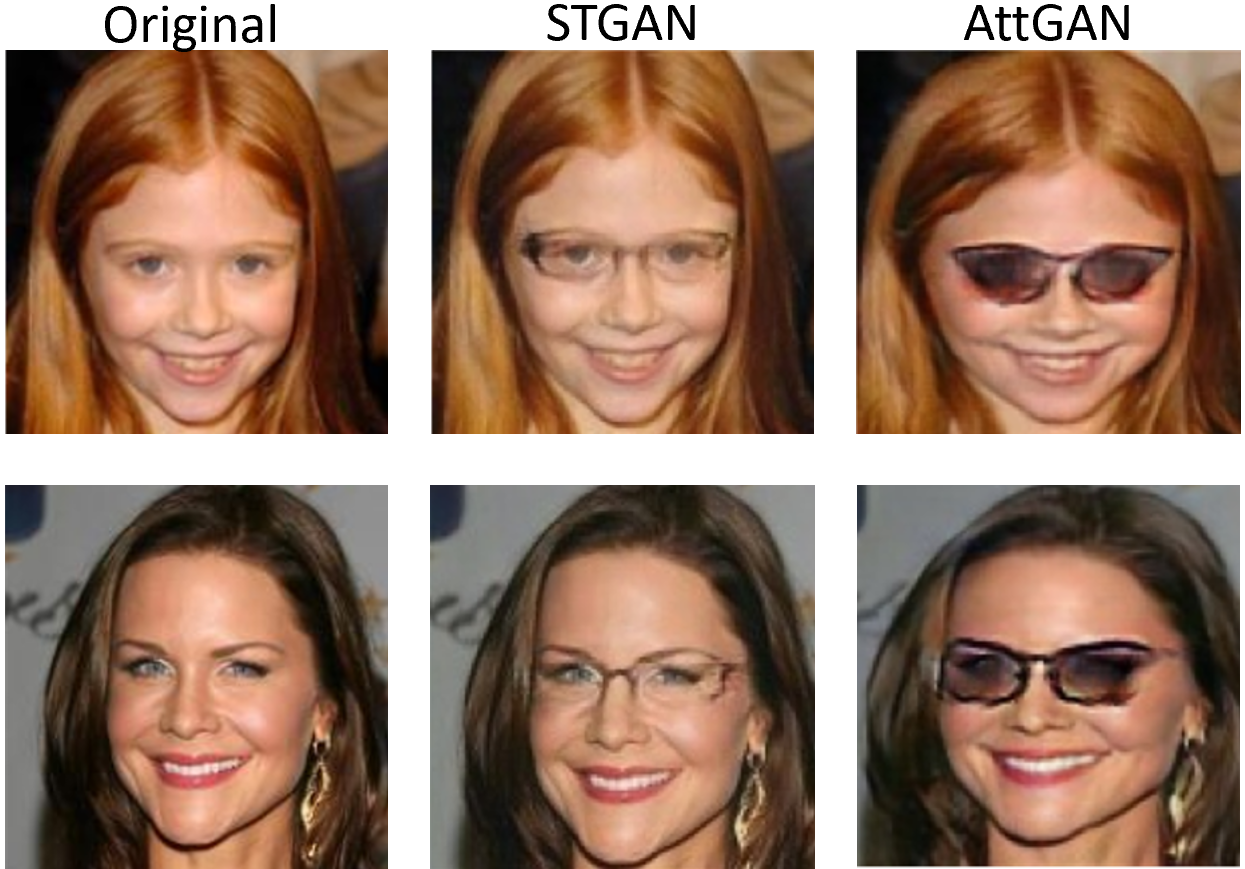}
    }  
    \subfloat[With Eyeglasses]
    {
    \centering
    \includegraphics[scale=0.25]{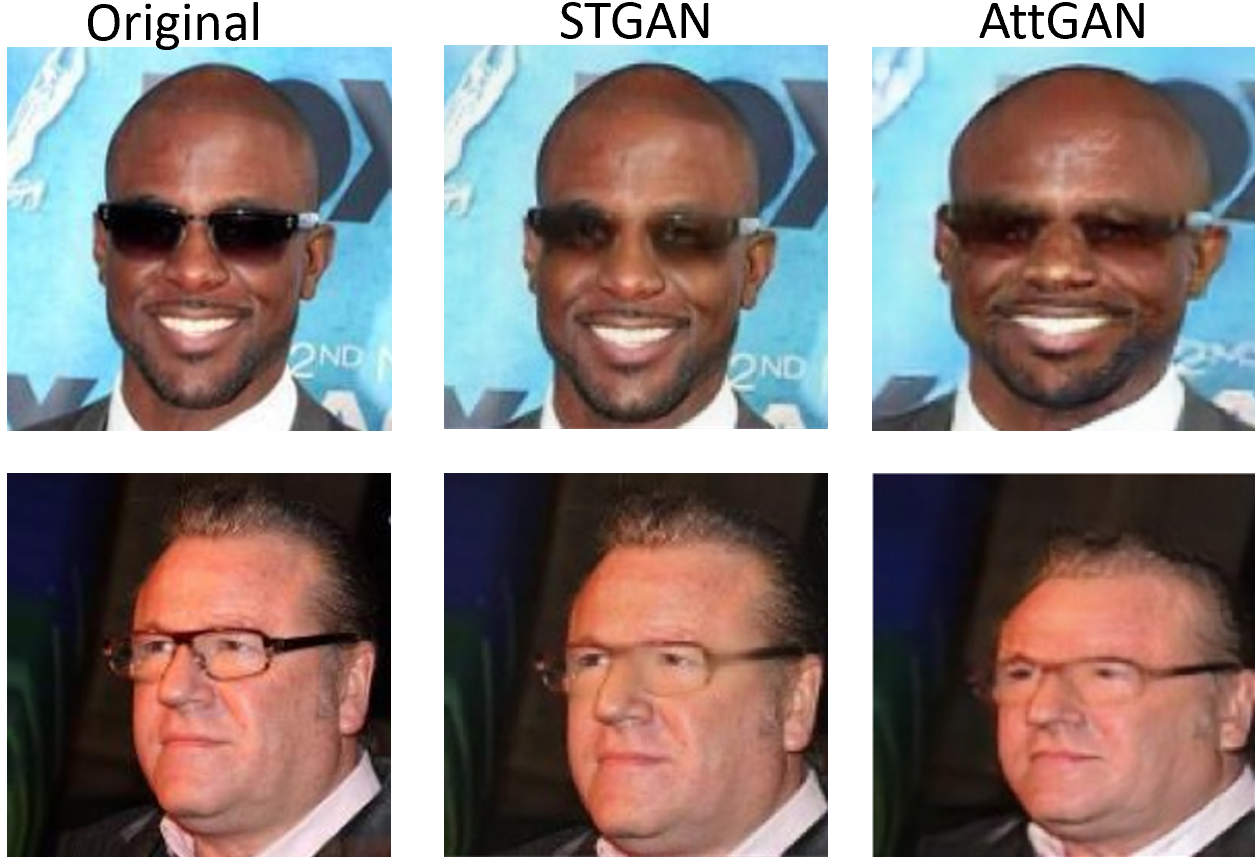}
    } 
    \caption{Examples depicting `Eyeglasses' attribute edited using STGAN and AttGAN, respectively, in original images (a) Without Eyeglasses and (b) With Eyeglasses. Note in images containing eyeglasses, the GANs fail to remove the glasses completely, instead they lighten the shades of the eyeglasses or the glass rims. STGAN outputs appear better for images with and without eyeglasses.}
    \label{fig:Eyeglasses}
\end{figure}

\begin{table}[h]
\centering
\caption{Face recognition performance: False Non-Match Rate (FNMR) at a False Match Rate \textbf{(FMR) = 0.01/0.1} for attribute manipulated images on the \textbf{LFW dataset}. The red colored cells correspond to the maximum degradation (if any) in the biometric recognition performance compared to the original in each column. Note \textit{all} of the attributes resulted in degradation. `Eyeglasses' caused the worst degradation in biometric recognition performance.}
\scalebox{0.935}{
\begin{tabular}{|l|l|l|} \hline
                       & \textbf{\begin{tabular}[c]{@{}l@{}}AttGAN-\\ ArcFace\end{tabular}} & \textbf{\begin{tabular}[c]{@{}l@{}}AttGAN-\\ VGGFace\end{tabular}} \\ \hline \hline
\textcolor{red}{Original}              & 0.17/0.11                                                                  &      0.39/0.17                                                                     \\ \hline
Bald                  &    0.39/0.16                                                             & 0.57/0.27                                                                                                                           \\
Bangs                 &     0.86/0.26                                                              & 0.71/0.39                                                                                                                           \\
Black\_Hair           &       0.33/0.14                                                            & 0.56/0.27                                                                                                                        \\
Blond\_Hair           &    0.36/0.14                                                               & 0.59/0.27                                                                                                                  \\
Brown\_hair           &    0.33/0.14                                                              & 0.59/0.31                                                                                                   \\
Bushy\_Eyebrows       &      0.50/0.20                                                             & 0.68/0.35                                                                                                                          \\
Eyeglasses            &      \cellcolor{red!25}0.99/0.43                                                             & \cellcolor{red!25}0.77/0.49                                                                                                                           \\
Male                  &   0.26/0.11                                                                & 0.47/0.23                                                                                                                               \\
Mouth\_Slightly\_Open &    0.30/0.12                                                               & 0.51/0.23                                                                     \\
Mustache              &   0.36/0.16                                                                & 0.56/0.28                                                                                                                              \\
No\_Beard             &     0.33/0.14                                                              & 0.55/0.26                                                                                                                            \\
Pale\_Skin            &       0.33/0.14                                                            & 0.56/0.26                                                                                                                                 \\
Young                 &    0.32/0.13                                                               & 0.55/0.26                                                                      \\ \hline                                                       
\end{tabular}}
\label{Tab3}
\end{table}

To investigate the possibility of \textbf{dataset-specific bias}, we selected 495 images from the Labeled Faces in-the-Wild (LFW) dataset~\cite{LFW} belonging to 100 individuals, and applied attribute editing using AttGAN. AttGAN resulted in the worst drop in performance on the CelebA dataset, so we employed it for this experiment. This is done to examine whether the impact of attribute editing on face recognition performance manifests across datasets. Next, we computed the face recognition performance between the original and the attribute-edited images using ArcFace and VGGFace comparators and observed the following results. `Eyeglasses' attribute editing resulted in the worst degradation in the performance by 83\% while using ArcFace and by 37\% while using VGGFace in terms of FNMR at FMR=0.01. It was followed by `Bangs' attribute resulting in second-highest degradation in the performance by 69\% while using ArcFace and by 32\% while using VGGFace in terms of FNMR at FMR=0.01. See Table~\ref{Tab3}.

To investigate the possibility of \textbf{GAN-specific artifacts}, we compared the `Reconstructed' images with the attribute edited images, specifically, the `Eyeglasses' and `Male' attribute edited images, to evaluate the face recognition performance. This is done to assess whether face recognition performance is affected solely by attribute editing or influenced by GAN-specific artifacts. Note that `Reconstructed' images are faithful reconstruction of `Original' images that demonstrate the fidelity of the GAN as an effective autoencoder. See Table~\ref{Tab4} for FNMR at FMR=0.01 and 0.1 for Reconstructed-Reconstructed, Reconstructed-Eyeglasses and Reconstructed-Male comparisons. We observed similar degradation in performance when comparing attribute edited images with GAN-reconstructed images by up to 47\% for `Eyeglasses' attribute and by up to 12\% for `Male' attribute in terms of FNMR at FMR=0.01. The results indicate that AttGAN has an overall weaker reconstruction and attribute editing capability than STGAN.

\textbf{\begin{table}[]
\centering
\caption{Face recognition performance: False Non-Match Rate (FNMR) at a False Match Rate \textbf{(FMR) = 0.01/0.1} for attribute manipulated images. We report the face recognition performance by comparing `Reconstructed' images with `Eyeglasses' attribute and `Male' attribute edited images.}
\begin{tabular}{|l|l|l|l|l|} \hline
              & \textbf{\begin{tabular}[c]{@{}l@{}}STGAN-\\ VGGFace\end{tabular}} & \textbf{\begin{tabular}[c]{@{}l@{}}STGAN-\\ ArcFace\end{tabular}} & \textbf{\begin{tabular}[c]{@{}l@{}}AttGAN-\\ VGGFace\end{tabular}} & \textbf{\begin{tabular}[c]{@{}l@{}}AttGAN-\\ ArcFace\end{tabular}} \\ \hline \hline
Reconstructed & 0.30/0.08                                                         & 0.21/0.10                                                         & 0.42/0.12                                                          & 0.49/0.20                                                          \\
Eyeglasses    & 0.38/0.12                                                         & 0.27/0.11                                                         & 0.50/0.19                                                          & 0.96/0.29                                                          \\
Male          & 0.42/0.16                                                         & 0.28/0.12                                                         & 0.47/0.20                                                          & 0.46/0.19   \\ \hline                                                      
\end{tabular}
\label{Tab4}
\end{table}
}

The implications of our findings are as follows. Bias due to naturally prevalent demographic factors in automated face recognition systems can be further aggravated when attributes are digitally modified. Digitally altering the sex cues (denoted by `Male' attribute) can be considered as manipulating a demographic attribute. It involves adding facial hair to images of female individuals and adding makeup to impart feminine appearance to images of male individuals. See Figure~\ref{fig:MaleAttrib}. These artificial manipulations affect face recognition performance. Surprisingly, altering a facial attribute like `Eyeglasses' caused an excessive degradation in face recognition performance. Modifying the `Eyeglasses' attribute involves adding glasses to individual face images where no eyeglasses are present and removing eyeglasses in the images where the individual is wearing one. See Figure~\ref{fig:Eyeglasses}. AttGAN struggles with addition as well as removal of eyeglasses from the images, and, instead, produces visually apparent artifacts that might be responsible for significant degradation in the biometric recognition performance. Removal of glasses is particularly hard: the GANs are only able to lighten the lens shades or the color of the eyeglass frames, but not completely remove the glasses. To check for any statistical variation between the results produced by AttGAN for the `Eyeglasses' attribute, we repeated the experiment with the original images five times. Each time, we followed the same procedure and executed basic attribute editing and not attribute sliding (sliding regulates the intensity of attribute modification). We obtained exactly identical results for each of the five runs, \textit{i.e.,} a decrease by 73\% in FNMR @FMR=0.01. On the LFW dataset, `Eyeglasses' attribute resulted in a decrease by up to 83\% in FNMR @FMR=0.01, while `Bangs', also a type of facial attribute, reduced the biometric recognition performance by up to 69\%. Therefore, we observed that `Eyeglasses' attribute editing reduced the biometric recognition performance considerably on both CelebA and LFW datasets. Although the exact reason responsible for significant degradation in face recognition performance caused by `Eyeglasses' attribute editing is unknown, we speculate that the attribute manipulations may produce severe changes in texture around the facial landmarks resulting in the drop in face recognition performance. 

\textbf{Therefore, our findings indicate that digitally modified attributes, both demographic and facial, can have a major impact on automated face recognition systems and can potentially introduce new biases that require further examination.}

\section{Conclusion} 
\label{Sec:Summ}
In this paper, we examined the impact of GAN-based attribute editing on face images in terms of face recognition performance. GAN-generated images are typically evaluated with respect to visual realism but their influence on biometric recognition is rarely analyzed. Therefore, we studied face recognition performance obtained using ArcFace and VGGFace after modifying thirteen attributes via AttGAN and STGAN on a total of $\sim$68,000 images belonging to 853 individuals from the CelebA dataset. Our findings indicated some interesting aspects: (i) Insertion or deletion of eyeglasses from a face image can significantly impair biometric recognition performance by up to 73\%. Digitally modifying the sex cues caused the second highest degradation in the performance by up to 22\%. (ii) There can be an artificial boost in the recognition accuracy by up to 7\% depending on the GAN used to modify the attributes and the deep learning-based face recognition network used to evaluate the biometric performance. Similar observations were reported when tested on a different dataset. Our findings indicate that attribute manipulations accomplished via GANs can significantly affect automated face recognition performance and need extensive analysis. Future work will focus on examining the effect of editing multiple attributes simultaneously, in a single face image, on face recognition.

\bibliographystyle{splncs04}
\balance
\bibliography{refs}
\end{document}